\documentclass{article}

\usepackage[preprint]{neurips_2024}

\usepackage[utf8]{inputenc} 
\usepackage[T1]{fontenc}    
\usepackage{url}            
\usepackage{booktabs}       
\usepackage{amsfonts}       
\usepackage{nicefrac}       
\usepackage{microtype}      
\usepackage{xcolor}         
\usepackage{color, colortbl}

\usepackage{amsmath}
\usepackage{amssymb}
\usepackage{mathtools}
\usepackage{amsthm}

\usepackage{tabu}
\usepackage{array, boldline, makecell}
\usepackage{graphicx}
\usepackage{caption}
\usepackage{subcaption}
\usepackage{multicol}
\usepackage{adjustbox}

\usepackage[ruled]{algorithm2e}
\usepackage{tikz}
\usepackage{changepage}

\usepackage{float}

\def\bea{\begin{eqnarray}} 
\def\eea{\end{eqnarray}}
\def\be{\begin{equation}} 
\def\ee{\end{equation}} 
\def\ba{\begin{array}}
\def\ea{\end{array}} 
\def\nn{\nonumber}

\def\gelu{\mathrm{GELU}}
\def\silu{\mathrm{SiLU}}
\def\relu{\mathrm{ReLU}}
\def\gelu{\mathrm{GELU}}
\def\elu{\mathrm{ELU}}
\def\lrelu{\mathrm{LeakyReLU}}

\newcommand{\inlinenote}[1]{
	\fcolorbox{orange}{yellow}{\parbox{\widthof{#1}}{#1}}
}

\newcommand{\inlinenotels}[1]{{\inlinenote{LS}}}

\usepackage[numbers]{natbib}

\usepackage[utf8]{inputenc} 
\usepackage[T1]{fontenc}    
\usepackage{url}            
\usepackage{booktabs}       
\usepackage{amsfonts}       
\usepackage{nicefrac}       
\usepackage{microtype}      
\usepackage{graphicx}

\usepackage{amsmath}
\usepackage{amssymb}
\usepackage{mathtools}
\usepackage{amsthm}
\usepackage{bm}

\usepackage{comment}

\usepackage{hyperref}       
\usepackage{xcolor}         

\title{Efficient Search for Customized Activation Functions with Gradient Descent}

\author{%
  Lukas Strack$^{1}$, Mahmoud Safari$^{1}$, Frank Hutter$^{2,1}$ \\
  $^{1}$Department of Computer Science, University of Freiburg\\
  $^{2}$ELLIS Institute Tübingen \\
  \texttt{\{strackl,safarim,fh\}@cs.uni-freiburg.de} \\
}

\begin{document}

\maketitle

\begin{abstract}
Different activation functions work best for different deep learning models. To exploit this, we leverage recent advancements in gradient-based search techniques for neural architectures to efficiently identify high-performing activation functions for a given application. We propose a fine-grained search cell that combines basic mathematical operations to model activation functions, allowing for the exploration of novel activations. Our approach enables the identification of specialized activations, leading to improved performance in every model we tried, from image classification to language models. Moreover, the identified activations exhibit strong transferability to larger models of the same type, as well as new datasets. Importantly, our automated process for creating customized activation functions is orders of magnitude more efficient than previous approaches. It can easily be applied on top of arbitrary deep learning pipelines and thus offers a promising practical avenue for enhancing deep learning architectures.
\end{abstract}

\section{Introduction} \label{intro}
Nonlinearities are an
indispensable component of any deep neural network. The design choice of these so-called activation functions has proven to crucially affect the training dynamics and final performance of neural networks. 

The rectified linear unit (ReLU) is
the most commonly used activation due to its simplicity and consistent performance across different tasks. However, it 
took several years of empirical research \citep{Hahnloser:2000aa, 5459469, nair2010rectified} before it 
was widely adopted by practitioners as an activation function in deep neural networks.

Despite the desirable properties of the ReLU, other alternatives
have been introduced \citep{maas2013rectifier, he2015delving, clevert2015fast, hendrycks2016gaussian, elfwing2018sigmoid, Misra2020MishAS}, each with their own theoretical or empirical justification, to address potential
issues associated with the ReLU, such as the dying ReLU problem~\citep{Trottier2016ParametricEL, Agarap2018DeepLU}. 
These alternative activations, which are mostly variations of ReLU,  
lead to performance improvements
in particular settings, although none is as widely adopted yet.

As evidenced by previous research, manually designing an activation function that suits a certain task is highly non-trivial and established choices (such as ReLU, GELU and SiLU) are made possibly at the cost of losing (optimal) performance. 
Automated search methods have been previously employed to learn activation functions (see Section \ref{related} for details), but existing methods require thousands of function evaluations and have thus not been adopted widely in practice. 
If it was possible to design a customized activation function for the problem at hand for the same cost as evaluating some standard alternatives (e.g., ReLU, GELU and SiLU) while yielding better performance, this would be quickly adopted by the community. That is the goal of our paper.

Our approach draws on recent developments in the 
rapidly growing field of Neural Architecture Search (NAS) with over a thousand papers in the last few years (see \cite{white2023neural} for a recent survey). 
NAS has mostly been limited to architectural choices, such as network depth or width in macro search spaces, or (choosing among) a pre-defined set of operations on the edges of a computational cell in cell-based search spaces, in all of which the activations are fixed. 
Recently, gradient-based one-shot methods \citep{liu2018darts, chen2020drnas, dong2019searching} have shown promise in efficiently optimizing architecture search spaces, reducing time costs by orders of magnitude compared to blackbox methods. Here, we adapt these NAS methods to mimic this success for searching activation functions by combining primitive mathematical operations.

We summarize our contributions as follows:
\begin{itemize}

\item We implement several key adjustments to modern gradient-based architecture search methods, tailoring them to search within the space of activations. This method is then integrated with a search space design of activations which is rich enough to accommodate novel activations, yet small enough to maintain search efficiency.

\item Within a wide range of image classification tasks, with ResNet and ViT architectures, as well as language modelling tasks with GPT, we demonstrate 
that using gradient-based one-shot search strategies we can discover from scratch specialized activations that improve a network's performance. Notably, our approach proves orders of magnitude more efficient compared to previous methods. 

\item Moreover, we investigate the transferability of the discovered activations to different models and datasets, and show that activation functions selected on a network/dataset
, are among the top-performing activations on similar but larger models, as well as on new datasets.
\end{itemize}

To facilitate reproducibility, we make our code available \href{https://github.com/automl/GRAFS}{here}.

\section{Related work} \label{related}

A line of research in automated activation function design utilizes gradient descent to learn ``adaptable activations'' during training together with network weights.
These works rely on a sufficiently general parameterization of activation functions that is capable of approximating a wide class of functions including most existing activations. \cite{agostinelli2014learning} use a general piecewise linear unit to approximate activations, while \cite{goyal2019learning} adopt a weighted sum of polynomial basis elements. Instead, \cite{molina2019pade} rely on the Pad\'{e} approximant (rational functions of polynomials) which shows better stability properties. Following \cite{agostinelli2014learning}, \cite{tavakoli2021splash} also adopt a piecewise linear approximation but introduce inductive bias to restrict the parameter space and provide a balance between simplicity and expressivity, hence simplifying optimization.

A separate approach \cite{ramachandran2017searching, basirat2018quest, bingham2020evolutionary, liu2020evolving, bingham2022discovering}, which is more in the spirit of NAS and further aligned with our current work, considers activation functions as hyper-parameters which are optimized in a search phase. The optimized function is then used as a fixed activation, possibly with learnable parameters, within a neural network. Contrary to gradient methods discussed previously, in this series of papers the activations within the search space are represented symbolically as combinations of basic functions. Moreover, they all utilize black-box optimization methods to explore the search space and thus require thousands of functions evaluations.

\cite{ramachandran2017searching} define the search space as a combination of basic unary and binary operations, and employ a search strategy previously developed for NAS \cite{DBLP:conf/iclr/ZophL17}. They utilize an RNN controller to sequentially predict different components of the activation function. The RNN controller is trained with reinforcement learning taking the validation accuracy of a proxy network/task with the candidate activation as the reward. With a combination of exhaustive and black-box search procedures, with a budget of 10\,000 function evaluations, they identify the Swish function as a high-performing activation that also generalizes across a variety of tasks.   

Along the same line, a number of subsequent works use evolutionary strategies to explore the space of activations. \cite{basirat2018quest} define the search space as consisting of separate pieces for negative and positive input, each of which is constructed from existing, well-known, activations, including Swish and two other activations, ELiSH and HardELiSH, introduced in the same paper, inspired by Swish. 
\cite{bingham2020evolutionary} apply evolution to a search space similar to the one of \cite{ramachandran2017searching}. \cite{liu2020evolving} search for both activation and normalization layers jointly as a single building block. 
The search space consists of a Directed Acyclic Graph (DAG) with basic mathematical functions (including unary and binary operations), as well as statistical moments on the nodes. More recently, \cite{bingham2022discovering} used evolutionary methods to search over a more flexible combination of unary and binary operations. The set of unary operations is slightly different from \cite{ramachandran2017searching} and includes existing high-performing activations, such as ReLU and Swish. As part of the evolutionary process, adaptable parameters are also randomly introduced in the activations which are then learned during training as any parametric activation. In a subsequent work,
AQuaSurF \cite{bingham2023efficient} introduced a surrogate representation by combining the Fisher information matrix eigenvalues and activation outputs through UMAP embeddings. 
This enabled a regression algorithm to search over this space efficiently, reducing the cost to a hundred function evaluations as opposed to thousands required by previous approaches.

The black-box nature of all these optimization methods makes them computationally demanding and impractical to apply to large search spaces and modern, costly, deep learning pipelines. In this work, we instead rely on gradient descent to explore the space of activation functions. We closely follow \cite{ramachandran2017searching}, \cite{bingham2020evolutionary}, and \cite{bingham2022discovering} to define the search space as combinations of low-level mathematical operations, as well as some existing activation functions. Contrary to previous gradient-based approaches, the search is performed in a bi-level fashion where the parameters of the activations are updated at the upper optimization level while the network weights are learned in the lower loop. This allows us to perform the optimization in the time it would require to evaluate only a few activation functions. The found activations can then be placed in the same or a different neural network which is trained from scratch.

\section{Methodology} \label{method}
We first describe our search space for activation functions, then discuss tools from gradient-based neural architecture search (NAS) we build on, and then discuss how we adapt them for effective gradient-based activation function search.

\subsection{The search space for activation functions}
\label{search-space}

Following \cite{ramachandran2017searching}, \cite{bingham2020evolutionary} and \cite{bingham2022discovering}, the space of activation functions is defined as a combination of unary and binary operations, which form a scalar function $f$, as shown in Figure \ref{fig:ss}. The unary and binary functions are chosen from a set of primitive mathematical operations, as listed in Figure \ref{fig:ss} (Left). We also include several existing activation functions as unary operations to enrich the search space further as in \cite{bingham2022discovering}.

The unary edges and binary vertices of the computational graph in Figure \ref{fig:ss} (Right) can take any of the corresponding operations from Figure \ref{fig:ss} (Left). In order to enable gradient-based optimization on this discrete space we continuously relax the space by assigning a weighted sum of all unary operations $\sum_{u} \upsilon^{(i,j)}_u u$ to the edge $(i,j)$ of the graph, and a weighted sum $\sum_{b} \beta^{(i)}_b b$ of binary operations to vertex $i$. Here the sums run over $u,b$ which denote respectively unary and binary operations in Figure \ref{fig:ss} (Left), and $\upsilon_u, \beta_b$ are the weights with which they appear in this sum. Both sets of coefficients are constrained to lie on a simplex $\sum_{u} \upsilon^{(i,j)}_u=\sum_{b} \beta^{(i)}_b=1$.

\begin{figure}[t] 
  \centering
  \begin{subfigure}{0.595\textwidth}
    \centering
 \resizebox{\columnwidth}{!}{$\displaystyle
\ba{ll|l}
\toprule
\multicolumn{2}{c}{\mathrm{\bf Unary}} & \multicolumn{1}{c}{\mathrm{\bf Binary}} \\ \hline
x 
& \sinh(x)
& x_1 + x_2  \\

-x 
& \tanh(x)
& x_1 - x_2 \nn \\

x^2 
& \mathrm{arcsinh}(x)
& x_1 x_2 \\

x^3 
& \mathrm{arctan}(x)
& \mathrm{max}(x_1, x_2) 
\\

\sqrt{x} 
& \mathrm{erf}(x)
& \mathrm{min}(x_1, x_2) \\

e^x  
& \mathrm{min}(0, x)
& \sigma(x_1)\, x_2 \\

|x| 
& \mathrm{max}(0, x) 
& \sigma(\gamma) x_1 + (1 - \sigma(\gamma)) x_2  \\

\gamma  
& \gelu(x) 
& L(x_1, x_2) \nn \\

\gamma x  
& \silu(x) 
& R(x_1, x_2)  \\

x+\gamma 
& \elu(x)
& \\

\sigma(x) 
& \lrelu(x)
& \\

\log(1+e^x)
& 
& \\
\bottomrule
\ea$}
\label{tab:ssl}
  \end{subfigure}
\begin{subfigure}{0.37\textwidth}
    \centering
    \begin{tikzpicture}[rounded corners=5pt,inner sep=2pt,node distance=2cm] 

    \node[,black] (Z) at (0,1) {$f(x)$};

    \node[draw,black,fill=brown!50, rounded corners=2pt, minimum width=10mm,minimum height=5mm] (A) at (0,0) {binary};
    \node[draw,brown,fill=orange!10!yellow!50!pink!20, rounded corners=2pt, minimum width=5mm,minimum height=5mm] (B) at (-0.5,-1) {unary};
    \node[draw,brown,fill=orange!10!yellow!50!pink!20, rounded corners=2pt, minimum width=5mm,minimum height=5mm] (C) at (0.5,-1) {unary};
    \node[draw,black,fill=brown!50, rounded corners=2pt, minimum width=10mm,minimum height=5mm] (D) at (-1,-2) {binary};

      \node[,black] (E) at (1,-2) {$x$};

    \draw[-,black,line width=1.5pt] (Z) -- (A) node[midway,above,sloped] {};

    \draw[-,black,line width=1.5pt] (A) -- (B) node[midway,above,sloped] {};
    \draw[-,black,line width=1.5pt] (A) -- (C) node[midway,above,sloped] {};

    \draw[-,black,line width=1.5pt] (B) -- (D) node[midway,above,sloped] {};
    \draw[-,black,line width=1.5pt] (C) -- (E) node[midway,above,sloped] {};

    \node[draw,black,fill=brown!50, rounded corners=2pt, minimum width=10mm,minimum height=5mm] (F) at (-1,-2) {binary};
    \node[draw,brown,fill=orange!10!yellow!50!pink!20, rounded corners=2pt, minimum width=5mm,minimum height=5mm] (G) at (-1.5,-3) {unary};
    \node[draw,brown,fill=orange!10!yellow!50!pink!20, rounded corners=2pt, minimum width=5mm,minimum height=5mm] (H) at (-0.5,-3) {unary};

    \node[,black] (I) at (-2,-4) {$x$};
    \node[,black] (J) at (0,-4) {$x$};

    \draw[-,black,line width=1.5pt] (F) -- (G) node[midway,above,sloped] {};
    \draw[-,black,line width=1.5pt] (F) -- (H) node[midway,above,sloped] {};

    \draw[-,black,line width=1.5pt] (G) -- (I) node[midway,above,sloped] {};
    \draw[-,black,line width=1.5pt] (H) -- (J) node[midway,above,sloped] {};
\end{tikzpicture}
\label{fig:ssr}
  \end{subfigure}
  \caption{(Left) set of unary and binary operations. $\gamma$ is a learnable parameter that is trained along with the activation parameters and becomes frozen after the search is completed. $\sigma(x)$ is the sigmoid function, and $L,R$ are the left and right projection operations. (Right) activation cell: combination of unary and binary operations}
\label{fig:ss}
\end{figure}
The computational cell in Figure \ref{fig:ss} (Right) is therefore a function of the \textit{activation parameters} $\upsilon, \beta$.
This will replace the original activation (ReLU for ResNet and GELU for ViT and GPT) within the network where the gradient-based search is carried out. The parameter $\gamma$ in Figure \ref{fig:ss} (Left) is a learnable parameter that is trained along with the activation parameters and becomes frozen after the search is completed.

\subsection{Tools from gradient-based neural architecture search} \label{nas}
We first review well-established gradient-based NAS methods, which will serve as a starting point for our gradient-based activation function search.

DARTS \cite{liu2018darts} was the first neural architecture search method that combined the weight-sharing idea \cite{pham2018efficient} with a continuous relaxation of architecture parameters,  allowing the use of gradient-descent to explore the architecture search space. 
This is carried out through bi-level optimization where gradient update steps are performed on continuous architectural parameters $\alpha$ in the outer loop, while model weights $w$ are updated in the inner loop:
\be
\ba{l}
\displaystyle \min_{\alpha}\, \mathcal{L}_{val}(w^*(\alpha), \alpha) \\[3mm]
\displaystyle s.t. \, w^*(\alpha) = \arg\!\min_{w}\, \mathcal{L}_{train}(w, \alpha)
\ea
\ee
In our specific problem of Section \ref{search-space} $\alpha$ will represent the collection of unary and binary parameters $(\upsilon, \beta)$. After the bi-level search phase is over, a final discretization step is then required to identify an architecture in the search space. The method is known to suffer from performance degradation at discretizaion \cite{zela2020understanding}.

In order to overcome the problem of large generalization errors and also encourage more exploration in the search space, DrNAS \cite{chen2020drnas} formulates the differentiable architecture search as a distribution learning problem where the architecture parameters $\alpha$ are sampled from a Dirichlet distribution $\alpha\sim Dir(\rho)$ with learnable parameters $\rho$.

Motivated by the success of
these methods in searching for top neural architectures, we employ similar search strategies to explore the space of activations. In particular, in this work, we opt to closely align with the distribution learning concept introduced in DrNAS (Algorithm \ref{alg:drnas}), based on its demonstrated effectiveness in architecture search and in our initial experiments.
However, given the slightly different nature of activation function spaces compared to those of neural architectures, this optimizer, at least in its original form, is not the best fit for discovering top performing activations. In the following subsection, we thus discuss how to modify it for searching the space of activations.

\subsection{Gradient-based activation function search}
\label{gradient-based-activation-search}

Given the similarity between the space of architectures and those of activation functions, described in the previous subsection, one may hope that existing architecture search techniques can be used out of the box to efficiently explore the space of activation functions. However, na\"ively applying gradient-based optimizers to activation search spaces simply fails. We hypothesize that this is why this approach does not exist in the literature yet for activation function search. In order to make gradient-based optimization work for such spaces, we now introduce a series of techniques to robustify the approach.

\paragraph{Warmstarting the search}
To robustify the search we introduce a short warm-starting phase during which the model weights are updated in the inner loop using the original activation, while the search cell is optimized in the outer loop. 
This ensures initializing the search with reasonable settings for both the network weights and the activation function parameters. 
After warm-starting the bi-level search continues, updating both model weights in the inner loop and activation parameters in the outer loop. 

\paragraph{Constraining unbounded operations}
Na\"ively applying gradient-based optimizers to activation search fails due to divergence of the search. This is caused by unbounded activation functions that lead to exploding gradients. To address this issue, 
we regularize the search space by constraining the unbounded operations in the search space. 
That is, operation outputs $y$ with magnitude beyond a threshold $|y|>\ell$ will be set to $y=\ell\,\mathrm{sign}(y)$. Here, we take $\ell=10$. After these two modifications, existing NAS methods can be run reliably on the space of activations, but we can improve performance further.

\paragraph{Progressive shrinking}
There are some fundamental differences between architecture spaces and those of activation functions. In particular, unlike architecture spaces, operations in the space of activations are nearly parameter free, as these are basic mathematical functions possibly with a few learnable parameters. Furthermore, different unary / binary elementary functions operate on different scales, making it challenging to rank their significance based on their coefficients.

Because of such inherent differences, it turns out that these methods do not perform well enough initially, at least to compete with existing activation baselines. 
Moreover the problem of performance drop at discretization, which is present in most NAS approaches, is more pronounced in the activation function space. To address these challenges we track activation parameters and at each epoch we drop a number of unary / binary operations corresponding to the lowest parameters (see Algorithm \ref{alg:drop}.). We choose the number of remaining operations to follow a logarithmic schedule\footnote{See Appendix \ref{app:schedule} for details.} such that at the final epoch we end up with a single unary(binary) operation on each edge(vertex), leading to a fully discretized activation. This {\it progressive shrinking} of the search cell not only improves efficacy of the approach but further expedites the search process.

\paragraph{DrNAS with variance reduction sampling} To optimize the activation cell we closely follow DrNAS, where a Dirichlet distribution $Dir(\rho)$ is assigned to each edge/vertex of the search cell and the concentration parameters $\rho$ are trained to minimize the expected validation loss.
At each iteration, DrNAS draws activation parameters from its Dirichlet distribution. While DrNAS by default uses a single fixed sample throughout the network, in our variant, in order to reduce the variance introduced by this sampling process, we draw independent samples for each activation cell within the network.  
Algorithm.\ref{alg:grafs} outlines the pseudocode for our {\bf GR}adient-based {\bf A}ctivation {\bf F}unction {\bf S}earch (GRAFS) approach. 

Besides architecture parameters, the activation cell includes a few learnable parameters represented by $\gamma$ in Figure \ref{fig:ss} (Left). These parameters are treated as part of activation parameters.
Upon completion of the bi-level search process, if operations involving learnable $\gamma$ variables are identified, their values will be fixed to their final learned values.

\begin{algorithm}[t]
    \caption{GRAFS}
    \label{alg:grafs}
    \SetKwInOut{Input}{Input}
    \Input{Shrinking schedule of the search cell $D_e$; Original activation function $\bar{a}$; Set of activation cells $\mathcal{A}$ that replace the original activation and their respective activation parameters $\alpha  = \{\alpha_a \vert \,\, \text{for} \,\, a \,\, \text{in} \,\, \mathcal{A}\}$; Total number of epochs $E$; Warm-starting epochs $E_0$}
    
    \textbf{Warm-starting:}\;
    \For{$e \leftarrow 1$ \KwTo $E_0$}{
        For all $a$ in $\mathcal{A}$ sample $\alpha_a\sim Dir(\rho)$\; Update distribution parameters $\rho$ by descending $\nabla_\rho\mathcal{L}_{valid}(w, \mathcal{A}(\alpha))$ \;
        Update weights $w$ by descending $\nabla_w \mathcal{L}_{train}(w, \bar{a})$\;
    }
    
    \textbf{Search:}\;
    \For{$e \leftarrow E_0$ \KwTo $E$}{
        DropOps($D_e$) \hfill $\triangleright$ see \hyperref[alg:drop]{Procedure \ref{alg:drop}}\;
        For all $a$ in $\mathcal{A}$ sample $\alpha_a \sim \text{Dir}(\rho)$\;
        Update distribution parameters $\rho$ by descending $\nabla_\rho \mathcal{L}_{valid}(w, \mathcal{A}(\alpha))$\;
        For all $a$ in $\mathcal{A}$ sample $\alpha_a \sim \text{Dir}(\rho)$\;
        Update weights $w$ by descending $\nabla_w \mathcal{L}_{train}(w, \mathcal{A}(\alpha))$\;
    }
\end{algorithm}

\begin{procedure}[t]
   \caption{DropOps($D$)}
   \label{alg:drop}
   \For{$i \leftarrow 1$ \KwTo $D$}{
       $O \gets$ edge or vertex with most operations left\;
       Drop operation in $O$ with lowest activation param\;
   }
\end{procedure}

\section{Experiments} \label{exps}
\subsection{Overview}

We explore high-performing activation functions across three distinct families of neural architectures: ResNet, ViT, and GPT. All examined network architectures in this study employ a single type of activation throughout the network. To conduct the search, the network's original activation function is globally replaced with the search cell in Figure \ref{fig:ss}. This activation cell is then optimized following the method outlined in Section \ref{gradient-based-activation-search}.

To assess our method's reliability, for each model, we repeat the search procedure with five different seeds, resulting in up to five distinct activation functions. In principle, this number could be less than five due to different searches converging to the same activation or known baseline activations. However, by retaining all distinct activations, even if they were very similar, as we did in this work, this did not occur in our experiments. The identified activation functions are evaluated on the networks/datasets they are searched on and subsequently also transferred to larger models of the same type and/or applied to new datasets.

For the evaluation of each discovered activation, we train the models with it for five seeds on the train set, and report test set performance (mean $\pm$ the standard error of the mean).

\subsection{Results}

\subsubsection{ResNet}

Residual networks (ResNets) were introduced in \cite{he2016deep} to mitigate the limitations of training deep neural networks and allow them to benefit from increased depth. They have since been the default in many image classification tasks.

In this section, our objective is to enhance the performance of ResNet20 trained on CIFAR10 by improving its activation functions. To achieve this, we replace the ReLU activations within ResNet20 with the search cell illustrated in Figure \ref{fig:ss}, and the exploration of the activation function space is carried out using the search strategy outlined in Section \ref{gradient-based-activation-search}. 

In all ResNet experiments, including the (inner loop) of the bi-level search and the (re)training of all models during evaluation, we utilized the PyTorch implementation provided in \cite{Idelbayev18a}. 

After five repetitions of the search process five distinct and new activation functions were identified. The explicit formulas are given as\footnote{The combinaition of two $\lrelu$s with default slope $10^{-2}$ in $F_{\mathrm{RN}}^{1}(x)$ is simply a $\lrelu$ with slope $10^{-4}$.}

{\setlength\arraycolsep{2pt}
\begin{eqnarray}
\label{eq:rn}
F_{\mathrm{RN}}^{1}(x) &=& 0.4739\,\lrelu(\lrelu(x)) + 0.5261\,\gelu(x) \nn \\ 
F_{\mathrm{RN}}^{2}(x) &=& 0.5163\,\lrelu(0.4945\,\relu(x) + 0.5055\, \gelu(x)) + 0.4837\,\gelu(x) \nn \\
F_{\mathrm{RN}}^{3}(x) &=& 0.4865\,\gelu(0.4873\,\relu(x) + 0.5127\, \gelu(x)) + 0.5135\,\gelu(x) \\
F_{\mathrm{RN}}^{4}(x) &=& 0.4756\,\relu(x) + 0.5244\,\gelu(x) \nn \\
F_{\mathrm{RN}}^{5}(x) &=& 0.4591\,\lrelu(0.5267\,\lrelu(x) + 0.4733\, \gelu(x)) + 0.5409\,\gelu(x) \nn 
\end{eqnarray}}%

and their functional forms are visualized in Appendix \ref{plots}. These five activations are then retrained from scratch on the training set and their performance is evaluated on the test set. The results are subsequently compared with those of existing baseline activation functions, including the original ReLU activation, as detailed in the left column of Table \ref{tab:rn}.
\begin{table}[t]
\centering
\resizebox{\textwidth}{!}{%
\begin{tabular}{lllllll}
\Xhline{1pt}
& \multicolumn{3}{c}{ResNet20} & \multicolumn{3}{c}{ResNet32} \\
\Xhline{1pt}
act.func & \multicolumn{1}{c}{CIFAR10} & \multicolumn{1}{c}{CIFAR100} & \multicolumn{1}{c}{SVHN Core} & \multicolumn{1}{c}{CIFAR10} & \multicolumn{1}{c}{CIFAR100} & \multicolumn{1}{c}{SVHN Core} \\
\toprule
$F_{\mathrm{RN}}^1$
& 91.87 $\pm$ 0.09
& 66.744 $\pm$ 0.157
& 95.797 $\pm$ 0.031
& 92.454 $\pm$ 0.271
& 68.256 $\pm$ 0.273
& 96.073 $\pm$ 0.019 \\ 
$F_{\mathrm{RN}}^2$
& 92.07 $\pm$ 0.109
& 66.946 $\pm$ 0.079
& 95.751 $\pm$ 0.038
& 92.708 $\pm$ 0.109
& 68.578 $\pm$ 0.2
& 96.13 $\pm$ 0.047 \\ 
$F_{\mathrm{RN}}^3$
& 91.838 $\pm$ 0.062
& 67.04 $\pm$ 0.166
& {\bf 95.87} $\pm$ 0.057
& 92.776 $\pm$ 0.087
& 68.084 $\pm$ 0.285
& {\bf 96.213} $\pm$ 0.073 \\ 
$F_{\mathrm{RN}}^4$
& {\bf 92.148} $\pm$ 0.1
& 66.916 $\pm$ 0.198
& 95.788 $\pm$ 0.04
& {\bf 92.864} $\pm$ 0.091
& 68.56 $\pm$ 0.234
& 96.098 $\pm$ 0.061 \\ 
$F_{\mathrm{RN}}^5$
& 92.008 $\pm$ 0.043
& 66.566 $\pm$ 0.122
& 95.76 $\pm$ 0.046
& 92.684 $\pm$ 0.039
& {\bf 68.636} $\pm$ 0.088
& 96.142 $\pm$ 0.027 \\ 
\midrule
$\silu$
& 91.902 $\pm$ 0.1
& 66.86 $\pm$ 0.091
& 95.658 $\pm$ 0.069
& 92.848 $\pm$ 0.077
& 68.528 $\pm$ 0.227
& 95.953 $\pm$ 0.043 \\ 
$\gelu$
& 92.034 $\pm$ 0.114
& 67.228 $\pm$ 0.094
& 95.828 $\pm$ 0.06
& 92.544 $\pm$ 0.061
& 68.474 $\pm$ 0.181
& 95.998 $\pm$ 0.032 \\ 
$\elu$
& 91.708 $\pm$ 0.06
& {\bf 67.42} $\pm$ 0.139
& 95.393 $\pm$ 0.026
& 92.23 $\pm$ 0.139
& 68.32 $\pm$ 0.183
& 95.586 $\pm$ 0.057 \\ 
$\lrelu$
& 91.656 $\pm$ 0.022
& 67.268 $\pm$ 0.217
& 95.681 $\pm$ 0.047
& 92.278 $\pm$ 0.097
& 68.276 $\pm$ 0.15
& 96.12 $\pm$ 0.041 \\ 
$\relu$
& 91.81 $\pm$ 0.063
& 66.862 $\pm$ 0.201
& 95.763 $\pm$ 0.059
& 92.494 $\pm$ 0.155
& 68.212 $\pm$ 0.307
& 96.079 $\pm$ 0.048 \\ 
\bottomrule
\end{tabular} 
} 
\vspace{1mm}
\caption{Test performance of activations found on ResNet20 / CIFAR10. Evaluations are on ResNet20 and ResNet32 / CIFAR10, CIFAR100, SVHN Core.} 
\label{tab:rn} 
\end{table} 
The remaining columns assess the generalization performance on CIFAR100 and SVHN Core datasets, as well as the larger model variant ResNet32, for all three datasets CIFAR10, CIFAR100 and SVHN Core. 

Table \ref{tab:rn} illustrates the effectiveness of our search method in identifying task-specific activation functions: On CIFAR10, two of the five activations surpass all baselines, and all five improve over the default ReLU activation. Furthermore, the newly discovered activations demonstrate strong transferability to larger models and new datasets, outperforming baselines in most cases. 

The overheads of search time over evaluation times on different models and datasets are shown in Table \ref{tab:search-eval-ratio-rn}, ranging from 2.2 to 4.1 function evaluations.
We note that the low ratios are partly due to the lower number of epochs used in the search process, and the aggressive pruning of the search cell at the early stages (See Appendix \ref{settings} for further details).
\begin{table}[H]
\footnotesize
\centering
\begin{tabular}{lccc}
\Xhline{1pt}
 & CIFAR10 
 & CIFAR100 
 & SVHN Core
 \\
 \toprule
 ResNet20 
 & 4.1
 & 4.1
 & 2.5  \\
 ResNet32 
 & 3.6
 & 3.6
 & 2.2 \\
\bottomrule
\end{tabular}
\vspace{1mm}
\caption{Search time to evaluation time ratios. Search is always on ResNet20 / CIFAR10.}
\label{tab:search-eval-ratio-rn}
\end{table}

\subsubsection{Vision Transformers}

After the success of the Transformer model \cite{vaswani2017attention} in natural language processing, Vision Transformers \cite{dosovitskiy2020image} based on the same self-attention mechanism have become increasingly popular in the vision domain.
In the original ViT model GELU has been the default activation function. Here we let the automated search discover the activation that is well-suited to the ViT architecture.

To avoid computational burden, we conduct the search on the ViT-Ti \cite{touvron2021training} model which is a light version of ViT. The specific version of this model, as well as a larger variant used for evaluation in this study, is adapted from the implementation provided by \cite{kentaroy47}, which we denote as ViT-tiny and ViT-small, respectively (See \ref{app:arch_setting} for details of the architectural choices). 

In the evaluation experiments of this section and in the inner loop of the search pipeline we utilized the GitHub repository \cite{kentaroy47}, but employed the TrivialAugment (TA) setup \cite{muller2021trivialaugment} as the augmentation method. TA simply applies a random augmentation with a random strength to each image, and has proved to achieve state-of-the-art on a variety of image classification tasks.  

Equation \ref{eq:vit} shows explicit formulas for the five novel activations found in the search process on ViT-tiny / CIFAR10. These activations are then evaluated and compared to baselines on ViT-tiny as well as the larger variant ViT-small on the three datasets CIFAR10, CIFAR100 and SVHN Core. 
The results, reported in Table \ref{tab:vit}, illustrate that all five activations outperform existing baselines on ViT-tiny / CIFAR10 providing high-performing customized activations for this task. Surprisingly, this pattern further extends to the datasets CIFAR100 and SVHN Core and the larger variant ViT-small.

The search overheads are collected in Table \ref{tab:search-eval-ratio-vit}, showing extremely small overheads of 0.16 to 0.26 function evaluations in this case. Note that overhead ratios smaller than one are feasible due to the reduced number of epochs employed during the search phase.
\begin{table}[H]
\footnotesize
\centering
\begin{tabular}{lccc}
\Xhline{1pt}
 & CIFAR10 
 & CIFAR100 
 & SVHN Core
 \\
 \toprule
 ViT-tiny 
& 0.26
& 0.25
& 0.17 \\ 
ViT-small 
& 0.23
& 0.24
& 0.16 \\
\bottomrule
\end{tabular}
\vspace{1mm}
\caption{Search time to evaluation time ratios. Search is always on ViT-tiny / CIFAR10.}
\label{tab:search-eval-ratio-vit}
\end{table}

To further evaluate the generalization capabilities of the discovered activations, we conduct additional experiments to assess their performance under an alternative training pipeline. 
Specifically, we utilized the timm library \cite{rw2019timm} and relied on the pipeline described by \cite{chen2021autoformer}, which has proved to be effective in training tiny versions of the ViT. 
Here, we incorporated the implementation for the tiny and small ViT models from \cite{rw2019timm}, which we denote as ViT-Ti and ViT-S, to distinguish them from the previously mentioned ViT-tiny and ViT-small models with distinct architectural settings (See \ref{app:arch_setting} for details and comparison).
Table \ref{tab:vit-ts} compares the test performance of our five activation functions with the five baseline activations for both ViT-Ti and ViT-S on CIFAR10. 
Remarkably, all the five discovered activations in Table \ref{tab:vit} consistently maintain their superior performance in this case. The longer training times in this case result in even smaller search overheads which are under $0.1$ in both cases. 
\begin{table}[t]
\centering
\resizebox{\textwidth}{!}{%
\begin{tabular}{lllllll}
\Xhline{1pt}
& \multicolumn{3}{c}{ViT-Tiny} & \multicolumn{3}{c}{ViT-Small} \\
\Xhline{1pt}
act.func & \multicolumn{1}{c}{CIFAR10} & \multicolumn{1}{c}{CIFAR100} & \multicolumn{1}{c}{SVHN} & \multicolumn{1}{c}{CIFAR10} & \multicolumn{1}{c}{CIFAR100} & \multicolumn{1}{c}{SVHN} \\
\toprule
$F_{\mathrm{ViT}}^1$
& 91.634 $\pm$ 0.188
& 69.94 $\pm$ 0.365
& 96.6 $\pm$ 0.02
& {\bf 94.06} $\pm$ 0.113
& 72.636 $\pm$ 0.255
& 97.163 $\pm$ 0.036 \\ 
$F_{\mathrm{ViT}}^2$
& 92.148 $\pm$ 0.067
& {\bf 70.462} $\pm$ 0.274
& 96.717 $\pm$ 0.026
& 94.046 $\pm$ 0.076
& 73.114 $\pm$ 0.289
& 97.141 $\pm$ 0.041 \\ 
$F_{\mathrm{ViT}}^3$
& 92.044 $\pm$ 0.135
& 70.074 $\pm$ 0.306
& 96.61 $\pm$ 0.053
& 93.912 $\pm$ 0.16
& 72.802 $\pm$ 0.222
& 97.16 $\pm$ 0.042 \\ 
$F_{\mathrm{ViT}}^4$
& 92.122 $\pm$ 0.135
& 70.142 $\pm$ 0.138
& 96.737 $\pm$ 0.026
& 93.91 $\pm$ 0.051
& 72.688 $\pm$ 0.253
& {\bf 97.19} $\pm$ 0.023 \\ 
$F_{\mathrm{ViT}}^5$
& {\bf 92.228} $\pm$ 0.195
& 70.232 $\pm$ 0.2
& {\bf 96.766} $\pm$ 0.017
& 93.764 $\pm$ 0.104
& {\bf 73.218} $\pm$ 0.173
& 97.138 $\pm$ 0.023 \\ 
\midrule
$\silu$
& 91.482 $\pm$ 0.214
& 68.802 $\pm$ 0.4
& 96.457 $\pm$ 0.062
& 93.412 $\pm$ 0.064
& 70.838 $\pm$ 0.439
& 97.078 $\pm$ 0.046 \\ 
$\gelu$
& 91.474 $\pm$ 0.115
& 68.374 $\pm$ 0.24
& 96.395 $\pm$ 0.055
& 93.282 $\pm$ 0.061
& 71.456 $\pm$ 0.144
& 97.018 $\pm$ 0.04 \\ 
$\elu$
& 90.888 $\pm$ 0.122
& 67.752 $\pm$ 0.298
& 96.365 $\pm$ 0.046
& 92.076 $\pm$ 0.132
& 67.462 $\pm$ 0.422
& 96.706 $\pm$ 0.033 \\ 
$\lrelu$
& 90.834 $\pm$ 0.136
& 68.148 $\pm$ 0.246
& 96.484 $\pm$ 0.048
& 92.906 $\pm$ 0.087
& 70.77 $\pm$ 0.248
& 96.941 $\pm$ 0.062 \\ 
$\relu$
& 91.05 $\pm$ 0.18
& 68.07 $\pm$ 0.102
& 96.477 $\pm$ 0.024
& 92.794 $\pm$ 0.068
& 70.282 $\pm$ 0.156
& 96.971 $\pm$ 0.046 \\ 
\bottomrule
\end{tabular} 
} 
\vspace{1mm}
\caption{Test performance of activations found on ViT-tiny / CIFAR10. Evaluations are on ViT-tiny and ViT-small with CIFAR10, CIFAR100 and SVHN Core datasets. All five discovered activations outperform baselines on all models and datasets.} 
\label{tab:vit} 
\end{table} 
\begin{table}[t]
\footnotesize
\centering
\begin{tabular}{lll}
\Xhline{1pt}
act.func & \multicolumn{1}{c}{ViT-Ti} & \multicolumn{1}{c}{ViT-S} \\
\toprule
$F_{\mathrm{ViT}}^1$
& 92.736 $\pm$ 0.107
& 94.264 $\pm$ 0.066 \\ 
$F_{\mathrm{ViT}}^2$
& 92.72 $\pm$ 0.103
& 94.162 $\pm$ 0.087 \\ 
$F_{\mathrm{ViT}}^3$
& 92.938 $\pm$ 0.122
& 94.292 $\pm$ 0.109 \\ 
$F_{\mathrm{ViT}}^4$
& 92.81 $\pm$ 0.081
& 94.28 $\pm$ 0.058 \\ 
$F_{\mathrm{ViT}}^5$
& {\bf 93.146} $\pm$ 0.186
& {\bf 94.336} $\pm$ 0.069 \\ 
\midrule
$\silu$
& 88.548 $\pm$ 0.083
& 93.056 $\pm$ 0.206 \\ 
$\gelu$
& 91.378 $\pm$ 0.089
& 94.098 $\pm$ 0.122 \\ 
$\elu$
& 82.878 $\pm$ 0.253
& 88.544 $\pm$ 0.246 \\ 
$\lrelu$
& 91.766 $\pm$ 0.217
& 93.886 $\pm$ 0.11 \\ 
$\relu$
& 91.666 $\pm$ 0.139
& 93.85 $\pm$ 0.136 \\ 
\bottomrule
\end{tabular} 
\vspace{1mm}
\caption{Test performance of activations found on ViT-tiny / CIFAR10. Evaluations are on ViT-Ti and ViT-S both on CIFAR10, with a training pipeline different from that used in the search. All discovered activations outperform baselines on both models.}
\label{tab:vit-ts}
\end{table} 
{\setlength\arraycolsep{2pt}
\bea
\label{eq:vit}
F_{\mathrm{ViT}}^{1}(x) &=& 0.6601 \,\gelu(\silu(x)\,\gelu(x)) + \, 0.3399 \,x^2 \nn \\
F_{\mathrm{ViT}}^{2}(x) &=& 0.7322 \,\silu(0.2822 \,x^2 + 0.7178 \,\gelu(x)) + 0.2678\,x^2 \nn \\
F_{\mathrm{ViT}}^{3}(x) &=& 0.7319 \,\gelu(\silu(x)\,\gelu(x)) + \, 0.2681 \,x^2 \\
F_{\mathrm{ViT}}^{4}(x) &=& 0.6778 \,\gelu(\silu(x)\,\gelu(x)) + \, 0.3222 \,x^2 \nn \\
F_{\mathrm{ViT}}^{5}(x) &=& 0.3139 \,x^2 + \, 0.5431 \,\gelu(x) \nn
\eea}%

\subsubsection{Generative pre-trained transformers}

To enhance the diversity of our experiments, we extend the evaluation of our approach to language modelling tasks. The best-established model in this domain is the Generative Pre-trained Transformer (GPT), which has recently achieved breakthrough performance. For the sake of simplicity, in this work we focus our analysis on Andrej Karpathy's nanoGPT\footnote{https://github.com/karpathy/nanoGPT}, a streamlined implementation of GPT-2 \cite{Radford2019LanguageMA}.

We optimize the activation within a down-scaled version of this architecture with 11M parameters featuring 3 layers, 3 heads and an embedding dimension of 192, which we denote as miniGPT. We employ the TinyStories \cite{eldan2023tinystories} dataset for training. 

As before, we repeat the search five times, warm-starting it with the default GELU activation. This results in the following five new activations: 
{\setlength\arraycolsep{2pt}
\bea
\label{eq:gpt}
F_{\mathrm{GPT}}^{1} &=& 0.4953\,\lrelu(x)\,\gelu(x) +  0.5047\,\relu(x) \nn \\
F_{\mathrm{GPT}}^{2} &=& (0.4689\,\gelu(x) +  0.5311)\relu(x) \nn\\
F_{\mathrm{GPT}}^{3} &=& (0.4662\,\sinh(x) +  0.5338)\gelu(x) \\
F_{\mathrm{GPT}}^{4} &=& 0.4781\,\relu(x)^2 + 0.5219\,\relu(x) \nn \\
F_{\mathrm{GPT}}^{5} &=& 0.4828\,\relu(x)^2 + 0.5172\,\relu(x) \nn
\eea}%
all of which demonstrate lower test losses compared to GELU, as detailed in the left column of Table \ref{tab:gpt}. 
As shown in the two right columns of this table, these improvements also transfer to two larger variants which we refer to as tinyGPT and smallGPT respectively. tinyGPT has 6 layers, 6 heads and an embedding dimension of 392, nearly tripling the size to 30M parameters, while smallGPT has 9 layers, 9 heads and an embedding dimension of 576 with 65M parameters.
\begin{table}[t] 
\footnotesize
\begin{center}
\begin{tabular}{llll}
\Xhline{1pt}
activ.func. & \multicolumn{1}{c}{miniGPT} & \multicolumn{1}{c}{tinyGPT} & \multicolumn{1}{c}{smallGPT} \\ 
\toprule
$F_{\mathrm{GPT}}^1$
& 1.933 $\pm$ 0.002
& 1.496 $\pm$ 0.002
& 1.321 $\pm$ 0.003 \\ 
$F_{\mathrm{GPT}}^2$
& 1.933 $\pm$ 0.001
& 1.495 $\pm$ 0.002
& 1.322 $\pm$ 0.002 \\ 
$F_{\mathrm{GPT}}^3$
& {\bf 1.921} $\pm$ 0.002
& {\bf 1.487} $\pm$ 0.002
& {\bf 1.317} $\pm$ 0.002 \\ 
$F_{\mathrm{GPT}}^4$
& 1.934 $\pm$ 0.003
& 1.495 $\pm$ 0.002
& 1.323 $\pm$ 0.002 \\ 
$F_{\mathrm{GPT}}^5$
& 1.933 $\pm$ 0.002
& 1.496 $\pm$ 0.002
& 1.323 $\pm$ 0.003 \\ 
\midrule
$\gelu$
& 1.943 $\pm$ 0.004
& 1.499 $\pm$ 0.002
& 1.325 $\pm$ 0.003 \\ 
\bottomrule
\end{tabular}
\end{center}
\caption{Activations identified by searching over miniGPT / TinyStories and evaluated on miniGPT, tinyGPT and smallGPT / TinyStories. Last row compares reults to original model with GELU activation. All discovered activations outperform GELU on all three models, with $F^3_{GPT}$ identified as the best activation.}
\label{tab:gpt}
\end{table}

The (asymptotic) $\relu(x)^2$ behaviour observed in $F^4_{GPT}$ and $F^5_{GPT}$ was previously identified in Primer \cite{so2021searching} through an evolutionary search over TensorFlow programs for Transformer language models, and was determined to be the most effective modification in the architecture.

The ratios of search time to evaluation time for all three models are reported in Table \ref{tab:search-eval-ratio-gpt}. Again, the extremely low ratios are due to lower number of iterations used during the search, and the initial aggressive shrinking of the activation cell.
\begin{table}[t]
\footnotesize
\centering
\begin{tabular}{lccc}
\Xhline{1pt}
 & miniGPT 
 & tinyGPT 
 & smallGPT 
 \\
\toprule
TinyStories 
& 1.1
& 0.7
& 0.5 \\ 
\bottomrule
\end{tabular}
\vspace{1mm}
\caption{Search time to evaluation time ratios. Search is always on miniGPT / TinyStories.}
\label{tab:search-eval-ratio-gpt}
\end{table}

\section{Conclusions} \label{conclusions}

We have adapted modern gradient-based architecture search techniques to explore the space of activation functions. Our work demonstrates that our proposed search strategy, when combined with a well-designed search space, can successfully identify activation functions tailored to specific deep learning models that surpass commonly-used alternatives. Furthermore, the discovered activation functions exhibit transferability to larger models of the same type, as well as new datasets, achieving high performance.

Most notably, the optimization is highly efficient, requiring very low overhead, up to only a few function evaluations in our case; this is 
in contrast to existing methods which require thousands of function evaluations. 
This makes it convenient for practitioners to employ these methods to automatically and efficiently design activation functions tailor-made for their deep learning architectures.

The method presented in this work aims to demonstrate the potential of gradient-based techniques in identifying top-performing activation functions, and as the first such work is not intended to represent the optimal pipeline for conducting such a search. While our approach, as is, may potentially already improve the strongest available models, we mostly see this work as opening the door for a host of possible follow-ups, 
such as improved search spaces and search methods, searching for activation functions with robust performance across workloads, or searching for activation functions with particularly strong scaling behavior to larger networks.
We hope that our work lays the ground for further research and exploration in this direction.

\clearpage

\section*{Acknowledements}

This research was funded by the Deutsche Forschungsgemeinschaft (DFG, German Research Foundation) under grant number 417962828. The authors acknowledge support by the state of Baden-Württemberg through bwHPC and the German Research Foundation (DFG) through grant INST 35/1597-1 FUGG. Frank Hutter is a Hector Endowed Fellow at the ELLIS Institute Tübingen.

\bibliography{main}

\begin{thebibliography}{10}

\bibitem{Agarap2018DeepLU}
Abien~Fred Agarap.
\newblock Deep learning using rectified linear units (relu).
\newblock {\em ArXiv}, abs/1803.08375, 2018.

\bibitem{agostinelli2014learning}
Forest Agostinelli, Matthew Hoffman, Peter Sadowski, and Pierre Baldi.
\newblock Learning activation functions to improve deep neural networks.
\newblock {\em arXiv preprint arXiv:1412.6830}, 2014.

\bibitem{basirat2018quest}
Mina Basirat and Peter~M Roth.
\newblock The quest for the golden activation function.
\newblock {\em arXiv preprint arXiv:1808.00783}, 2018.

\bibitem{bingham2020evolutionary}
Garrett Bingham, William Macke, and Risto Miikkulainen.
\newblock Evolutionary optimization of deep learning activation functions.
\newblock In {\em Proceedings of the 2020 Genetic and Evolutionary Computation
  Conference}, pages 289--296, 2020.

\bibitem{bingham2022discovering}
Garrett Bingham and Risto Miikkulainen.
\newblock Discovering parametric activation functions.
\newblock {\em Neural Networks}, 148:48--65, 2022.

\bibitem{bingham2023efficient}
Garrett Bingham and Risto Miikkulainen.
\newblock Efficient activation function optimization through surrogate
  modeling.
\newblock {\em arXiv preprint arXiv:2301.05785}, 2023.

\bibitem{chen2021autoformer}
Minghao Chen, Houwen Peng, Jianlong Fu, and Haibin Ling.
\newblock Autoformer: Searching transformers for visual recognition.
\newblock In {\em Proceedings of the IEEE/CVF international conference on
  computer vision}, pages 12270--12280, 2021.

\bibitem{chen2020drnas}
Xiangning Chen, Ruochen Wang, Minhao Cheng, Xiaocheng Tang, and Cho-Jui Hsieh.
\newblock Drnas: Dirichlet neural architecture search.
\newblock In {\em International Conference on Learning Representations}, 2020.

\bibitem{clevert2015fast}
Djork-Arn{\'e} Clevert, Thomas Unterthiner, and Sepp Hochreiter.
\newblock Fast and accurate deep network learning by exponential linear units
  (elus).
\newblock {\em arXiv preprint arXiv:1511.07289}, 2015.

\bibitem{dong2019searching}
Xuanyi Dong and Yi~Yang.
\newblock Searching for a robust neural architecture in four gpu hours.
\newblock In {\em Proceedings of the IEEE/CVF Conference on Computer Vision and
  Pattern Recognition}, pages 1761--1770, 2019.

\bibitem{dosovitskiy2020image}
Alexey Dosovitskiy, Lucas Beyer, Alexander Kolesnikov, Dirk Weissenborn,
  Xiaohua Zhai, Thomas Unterthiner, Mostafa Dehghani, Matthias Minderer, Georg
  Heigold, Sylvain Gelly, et~al.
\newblock An image is worth 16x16 words: Transformers for image recognition at
  scale.
\newblock In {\em International Conference on Learning Representations}, 2020.

\bibitem{eldan2023tinystories}
Ronen Eldan and Yuanzhi Li.
\newblock Tinystories: How small can language models be and still speak
  coherent english?
\newblock {\em arXiv preprint arXiv:2305.07759}, 2023.

\bibitem{elfwing2018sigmoid}
Stefan Elfwing, Eiji Uchibe, and Kenji Doya.
\newblock Sigmoid-weighted linear units for neural network function
  approximation in reinforcement learning.
\newblock {\em Neural Networks}, 107:3--11, 2018.

\bibitem{goyal2019learning}
Mohit Goyal, Rajan Goyal, and Brejesh Lall.
\newblock Learning activation functions: A new paradigm for understanding
  neural networks.
\newblock {\em arXiv preprint arXiv:1906.09529}, 2019.

\bibitem{Hahnloser:2000aa}
Richard H.~R. Hahnloser, Rahul Sarpeshkar, Misha~A. Mahowald, Rodney~J.
  Douglas, and H.~Sebastian Seung.
\newblock Digital selection and analogue amplification coexist in a
  cortex-inspired silicon circuit.
\newblock {\em Nature}, 405(6789):947--951, 2000.

\bibitem{he2015delving}
Kaiming He, Xiangyu Zhang, Shaoqing Ren, and Jian Sun.
\newblock Delving deep into rectifiers: Surpassing human-level performance on
  imagenet classification.
\newblock In {\em Proceedings of the IEEE international conference on computer
  vision}, pages 1026--1034, 2015.

\bibitem{he2016deep}
Kaiming He, Xiangyu Zhang, Shaoqing Ren, and Jian Sun.
\newblock Deep residual learning for image recognition.
\newblock In {\em Proceedings of the IEEE conference on computer vision and
  pattern recognition}, pages 770--778, 2016.

\bibitem{hendrycks2016gaussian}
Dan Hendrycks and Kevin Gimpel.
\newblock Gaussian error linear units (gelus).
\newblock {\em arXiv preprint arXiv:1606.08415}, 2016.

\bibitem{Idelbayev18a}
Yerlan Idelbayev.
\newblock Proper {ResNet} implementation for {CIFAR10/CIFAR100} in {PyTorch}.
\newblock \url{https://github.com/akamaster/pytorch_resnet_cifar10}.
\newblock Accessed: 20xx-xx-xx.

\bibitem{5459469}
Kevin Jarrett, Koray Kavukcuoglu, Marc'Aurelio Ranzato, and Yann LeCun.
\newblock What is the best multi-stage architecture for object recognition?
\newblock In {\em 2009 IEEE 12th International Conference on Computer Vision},
  pages 2146--2153, 2009.

\bibitem{liu2020evolving}
Hanxiao Liu, Andy Brock, Karen Simonyan, and Quoc Le.
\newblock Evolving normalization-activation layers.
\newblock {\em Advances in Neural Information Processing Systems},
  33:13539--13550, 2020.

\bibitem{liu2018darts}
Hanxiao Liu, Karen Simonyan, and Yiming Yang.
\newblock Darts: Differentiable architecture search.
\newblock In {\em International Conference on Learning Representations}, 2018.

\bibitem{maas2013rectifier}
Andrew~L Maas, Awni~Y Hannun, Andrew~Y Ng, et~al.
\newblock Rectifier nonlinearities improve neural network acoustic models.
\newblock In {\em Icml}, 2013.

\bibitem{Misra2020MishAS}
Diganta Misra.
\newblock Mish: A self regularized non-monotonic activation function.
\newblock In {\em British Machine Vision Conference}, 2020.

\bibitem{molina2019pade}
Alejandro Molina, Patrick Schramowski, and Kristian Kersting.
\newblock Pad{\'e} activation units: End-to-end learning of flexible activation
  functions in deep networks.
\newblock In {\em International Conference on Learning Representations}, 2019.

\bibitem{muller2021trivialaugment}
Samuel~G M{\"u}ller and Frank Hutter.
\newblock Trivialaugment: Tuning-free yet state-of-the-art data augmentation.
\newblock In {\em Proceedings of the IEEE/CVF International Conference on
  Computer Vision}, pages 774--782, 2021.

\bibitem{nair2010rectified}
Vinod Nair and Geoffrey~E Hinton.
\newblock Rectified linear units improve restricted boltzmann machines.
\newblock In {\em Icml}, 2010.

\bibitem{pham2018efficient}
Hieu Pham, Melody Guan, Barret Zoph, Quoc Le, and Jeff Dean.
\newblock Efficient neural architecture search via parameters sharing.
\newblock In {\em International conference on machine learning}, pages
  4095--4104. PMLR, 2018.

\bibitem{Radford2019LanguageMA}
Alec Radford, Jeff Wu, Rewon Child, David Luan, Dario Amodei, and Ilya
  Sutskever.
\newblock Language models are unsupervised multitask learners.
\newblock 2019.

\bibitem{ramachandran2017searching}
Prajit Ramachandran, Barret Zoph, and Quoc~V Le.
\newblock Searching for activation functions.
\newblock {\em arXiv preprint arXiv:1710.05941}, 2017.

\bibitem{so2021searching}
David So, Wojciech Ma{\'n}ke, Hanxiao Liu, Zihang Dai, Noam Shazeer, and Quoc~V
  Le.
\newblock Searching for efficient transformers for language modeling.
\newblock {\em Advances in neural information processing systems},
  34:6010--6022, 2021.

\bibitem{tavakoli2021splash}
Mohammadamin Tavakoli, Forest Agostinelli, and Pierre Baldi.
\newblock Splash: Learnable activation functions for improving accuracy and
  adversarial robustness.
\newblock {\em Neural Networks}, 140:1--12, 2021.

\bibitem{touvron2021training}
Hugo Touvron, Matthieu Cord, Matthijs Douze, Francisco Massa, Alexandre
  Sablayrolles, and Herv{\'e} J{\'e}gou.
\newblock Training data-efficient image transformers \& distillation through
  attention.
\newblock In {\em International Conference on Machine Learning}, pages
  10347--10357. PMLR, 2021.

\bibitem{Trottier2016ParametricEL}
Ludovic Trottier, Philippe Gigu{\`e}re, and Brahim Chaib-draa.
\newblock Parametric exponential linear unit for deep convolutional neural
  networks.
\newblock {\em 2017 16th IEEE International Conference on Machine Learning and
  Applications (ICMLA)}, pages 207--214, 2016.

\bibitem{vaswani2017attention}
Ashish Vaswani, Noam Shazeer, Niki Parmar, Jakob Uszkoreit, Llion Jones,
  Aidan~N Gomez, {\L}ukasz Kaiser, and Illia Polosukhin.
\newblock Attention is all you need.
\newblock {\em Advances in neural information processing systems}, 30, 2017.

\bibitem{white2023neural}
Colin White, Mahmoud Safari, Rhea Sukthanker, Binxin Ru, Thomas Elsken, Arber
  Zela, Debadeepta Dey, and Frank Hutter.
\newblock Neural architecture search: Insights from 1000 papers.
\newblock {\em arXiv preprint arXiv:2301.08727}, 2023.

\bibitem{rw2019timm}
Ross Wightman.
\newblock Pytorch image models.
\newblock \url{https://github.com/rwightman/pytorch-image-models}, 2019.

\bibitem{kentaroy47}
Kentaro Yoshioka.
\newblock \url{https://github.com/kentaroy47/vision-transformers-cifar10}.

\bibitem{zela2020understanding}
A~Zela, T~Elsken, Tonmoy Saikia, Yassine Marrakchi, Thomas Brox, and F~Hutter.
\newblock Understanding and robustifying differentiable architecture search.
\newblock In {\em International Conference on Learning Representations (ICLR)},
  2020.

\bibitem{DBLP:conf/iclr/ZophL17}
Barret Zoph and Quoc~V. Le.
\newblock Neural architecture search with reinforcement learning.
\newblock In {\em 5th International Conference on Learning Representations,
  {ICLR} 2017, Toulon, France, April 24-26, 2017, Conference Track
  Proceedings}. OpenReview.net, 2017.

\end{thebibliography}
\bibliographystyle{plain}


\newpage
\appendix

\section{Dirichlet Neural Architecture Search}
\label{app:drnas}
For completeness, in this section we present the pseudocode for DrNAS for neural architecture search. In \cite{chen2020drnas} an explicit regularizer term $\lambda d(\rho, \hat \rho)$ (specifically L2 norm) appears with coefficient $\lambda$ in the validation loss which enforces the distribution parameters $\rho$ to stay close to an anchor $\hat \rho=1$, and encourage exploration. Here and in Algorithm.\ref{alg:grafs} we omit this regularizer term for simplicity of notation, but it is important to note that an equivalent effect is achieved by using a nonzero weight decay in the optimizer.
\begin{algorithm}[H]
\footnotesize
\caption{DrNAS - Dirichlet Neural Architecture Search}
\label{alg:drnas}
 \SetKwInOut{Input}{Input}
    \SetKwInOut{Output}{Output}
    \Input{One-shot model with Initialized weights $w$; Dirichlet distribution parameters $\rho$; Anchor $\hat \rho=1$, anchor regularizer parameter $\lambda$, and metric $d$}
		\While{\textit{not converged}}{
            1. Sample architecture parameters $\alpha \sim Dir(\rho)$\; 
			2. Update distribution parameters $\rho$ by descending  $\nabla_\rho\mathcal{L}_{valid}(w, \alpha)$\;
			3. Sample architecture parameters $\alpha \sim Dir(\rho)$\;
   4. Update weights $w$ by descending $\nabla_w \mathcal{L}_{train}(w, \alpha)$}
		\textbf{Return:} Derive the final discretized architecture based on argmax of learned $\rho$
\end{algorithm}

\section{Shrinking schedule}
\label{app:schedule}
\begin{figure}[h]
\centering
\includegraphics[width=0.6\linewidth]{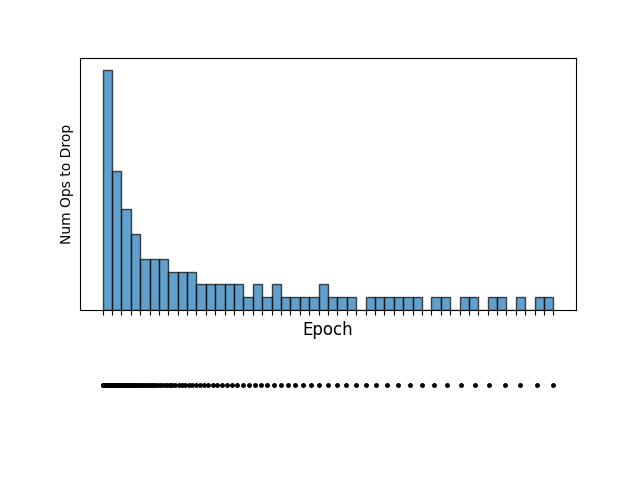}
\vspace{-5mm}
\caption{(Bottom) Log-scaled distribution of epochs at which operations are dropped. (Top) Histogram determines number of operations to drop per epoch.}
\label{fig:drop_ops}
\end{figure}

In Algorithm \ref{alg:grafs} the shrinking schedule $D_e$ denotes the number of operations to be dropped at epoch $e$ during the search phase. In this work we adopt a log schedule for $D_e$. Specifically, given the initial (total) number of operations in the activation cell $D=4\times 23 + 2\times 9=110$, $D-6$ operations have to be dropped in order to reach a fully discretized architecture with 6 operations. $D-6$ points are then distributed with a log spacing among the epochs, starting from epoch $e=S$, at which shrinking begins, and the final epoch $e=E$. These points are then binned into unit intervals, determining the number of operations to drop at each epoch (see Fig.\ref{fig:drop_ops} for a visualization). In this work we always start shrinking at twice the warm-starting epoch $S=2E_0$. 

\section{Architectural parameters}
\label{app:arch_setting}

Given the many versions of the ViT and GPT architectures, to avoid ambiguity, we present here the architectural parameters of the models we have used in this work. The two models ViT-Ti and ViT-S are simply the models \texttt{vit\_tiny\_patch16\_224} and \texttt{vit\_small\_patch16\_224} from \citep{rw2019timm}.
\begin{table}[h]
\centering
\begin{tabular}{lllll}
\Xhline{1pt}
 & \multicolumn{1}{c}{ViT-Ti}
 & \multicolumn{1}{c}{ViT-S}
 & \multicolumn{1}{c}{ViT-tiny}
 & \multicolumn{1}{c}{ViT-small}
 \\
 \toprule
embed\_dim  & 192 & 384 & 512 & 512
\\
depth & 12 & 12 & 4 & 6
\\
num\_heads & 3 & 6 & 6 & 8
\\
mlp\_dim & 768 & 1536 & 256 & 512
\\
patch\_size & 16 & 16 & 4 & 4
\\
img\_size & 224 & 224 & 32 & 32
\\
\bottomrule
\end{tabular}
\vspace{1mm}
\caption{Architectural parameters for ViT-Ti, ViT-S, ViT-tiny and ViT-small.}
\label{tab:vit_arch}
\end{table}
\begin{table}[H]
    \centering
    \begin{tabular}{lccc}
\Xhline{1pt}
        & miniGPT & tinyGPT & smallGPT \\ \toprule
        n\_layers & 3 & 6 & 9 \\
        n\_heads & 3 & 6 & 9 \\
        n\_embd & 192 & 384 & 576\\ \hline
    \end{tabular}
    \vspace{1mm}
    \caption{Architectural parameters defining miniGPT, tinyGPT and smallGPT.}
     \label{tab:gpt_arch}
\end{table}

\section{Experimental settings}
\label{settings}
In this section we provide the the details for the search and evaluation pipelines of the image classification and language modelling experiments. 

In Tables \ref{tab:rn_search}, \ref{tab:vit_search}, \ref{tab:gpt_search}, quantities above the separating line belong to the inner optimization, while those below the line are related to the outer loop.

All the search and evaluation experiments have been done on a single NVIDIA A40 GPU, except experiments on ViT-Ti which have been done on a single GeForce RTX 2080 Ti GPU.

\subsection{ResNet experiments}

\begin{table}[H]
    \centering
    \begin{tabular}{ll}
    \Xhline{1pt}
        \multicolumn{2}{c}{Search - ResNet20} \\
        \toprule
        Dataset & CIFAR10 \\
        Search epochs & 50 \\
        Batch size & 32 \\
        Gradient accumulation steps & 16 \\ 
        Optimizer & SGD(lr=0.1, momentum=0.9, weight decay=1e-4)\\
        \midrule
        Train-val split & 0.75 \\
        Arch optimizer & Adam(lr=0.0006, betas=(0.5, 0.999)) \\
        Warmstart epoch & 1 \\
        Start shrinking epoch & 2 \\
    \Xhline{1pt} \\
    \end{tabular}
    \caption{Hyperparameter settings for the bi-level search process on ResNet20 / CIFAR10.}
    \label{tab:rn_search}
\end{table}

\begin{table}[H]
    \centering
    \begin{tabular}{ll}
    \Xhline{1pt}
        \multicolumn{2}{c}{Evaluation - ResNet20, ResNet32} \\
    \toprule
        Dataset & CIFAR10, CIFAR100, SVHN Core \\
        Epochs & 200 \\
        Batch size & 128 \\ 
        Optimizer & SGD(lr=0.1, momentum=0.9, weight decay=1e-4) \\
        Learning rate & MultiStepLR with milestones=[100, 150] and gamma=0.1 \\ 
        \Xhline{1pt} \\
    \end{tabular}
    \caption{Hyperparameter settings for the evaluation process on ResNet20, ResNet32.}
    \label{tab:rn_eval}
\end{table}

\subsection{ViT experiments}

\begin{table}[H]
    \centering
    \begin{tabular}{ll}
    \Xhline{1pt}
        \multicolumn{2}{c}{Search - ViT-tiny} \\
        \toprule
        Dataset & CIFAR10 \\
        Augmentation & TrivialAugment \\
        Search epochs & 50 \\
        Batch size & 128 \\
        Gradient accumulation\_steps & 4 \\
        Optimizer & Adam(lr=0.001, betas=(0.9, 0.999))\\
        Learning rate & cosine annealing from 0.001 to zero \\ \midrule
        Train-val split & 0.75 \\
        Arch optimizer & Adam(lr=0.001, betas=(0.5, 0.999)) \\
        Warmstart epoch & 1 \\
        Start shrinking epoch & 2 \\
    \Xhline{1pt} \\
    \end{tabular}
    \caption{Hyperparameter settings for the bi-level search process on ViT-tiny / CIFAR10.}
    \label{tab:vit_search}
\end{table}

\begin{table}[H]
    \centering
    \begin{tabular}{ll}
    \Xhline{1pt}
        \multicolumn{2}{c}{Evaluation - ViT-tiny, ViT-small} \\
    \toprule
        Dataset & CIFAR10, CIFAR100, SVHN Core \\
        Augmentation & TrivialAugment \\
        Epochs & 500 \\
        Batch size & 512 \\
        Optimizer & Adam(lr=0.001, betas=(0.9, 0.999)) \\
        Learning rate & cosine annealing from $lr=10^{-4}$ to zero \\ 
            \Xhline{1pt} \\
    \end{tabular}
    \caption{Hyperparameter settings for the evaluation process on ViT-tiny, ViT-small.}
    \label{tab:vit_eval}
\end{table}

\subsection{GPT experiments}

\begin{table}[H]
    \centering
    \begin{tabular}{ll}
    \Xhline{1pt}
        \multicolumn{2}{c}{Search - miniGPT} \\
    \toprule
        Dataset & TinyStories \\
        Eval interval & 100 \\
        Max iters & 1000 \\
        Batch size & 4 \\
        Gradient accumulation\_steps & 40 \\ \midrule
        Train-val split & 0.75 \\
        Arch optimizer & Adam(lr=1e-3, betas=(0.5, 0.999))\\
        Warmstart iterations & 100 \\
        Start shrinking iteration & 200 \\
         \Xhline{1pt} \\
    \end{tabular}
    \caption{Hyperparameter settings for the bi-level search process on miniGPT.}
    \label{tab:gpt_search}
\end{table}

\begin{table}[H]
    \centering
    \begin{tabular}{ll}
   \Xhline{1pt}
        \multicolumn{2}{c}{Evaluation - miniGPT, tinyGPT, smallGPT} \\
    \toprule
        Dataset & TinyStories \\
        Compile & False \\
        Max iters & 10000 \\
        Batch size & 16 \\
        Gradient accumulation\_steps & 40 \\
        Optimizer & AdamW(lr=6e-4, weight\_decay=1e-1, betas=(0.9, 0.95))\\
        Learning rate & 100 linear warmup iters from $lr=0$ to $lr=6\!\times\!10^{-4}$ \\
        & then cosine annealing to $lr=6\!\times\!10^{-5}$ \\ 
                 \Xhline{1pt} \\
    \end{tabular}
    \caption{Hyperparameter settings for the evaluation process on miniGPT, tinyGPT and smallGPT.}
    \label{tab:gpt_eval}
\end{table}

\section{Activation function plots}
\label{plots}

\begin{figure}[H]
\centering
\begin{subfigure}{0.32\textwidth}
\centering
	\includegraphics[width=\linewidth]{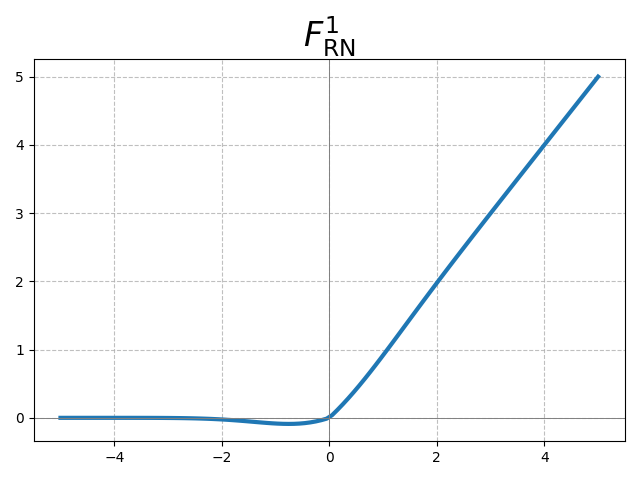}
\end{subfigure}
\begin{subfigure}{0.32\textwidth}
\centering
	\includegraphics[width=\linewidth]{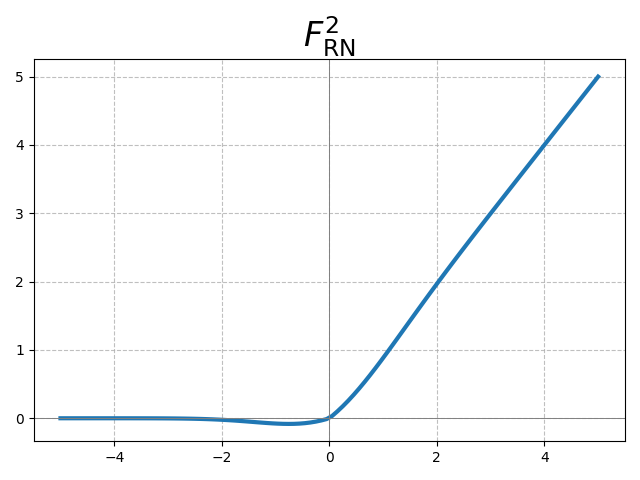}
\end{subfigure}
\begin{subfigure}{0.32\textwidth}
\centering
	\includegraphics[width=\linewidth]{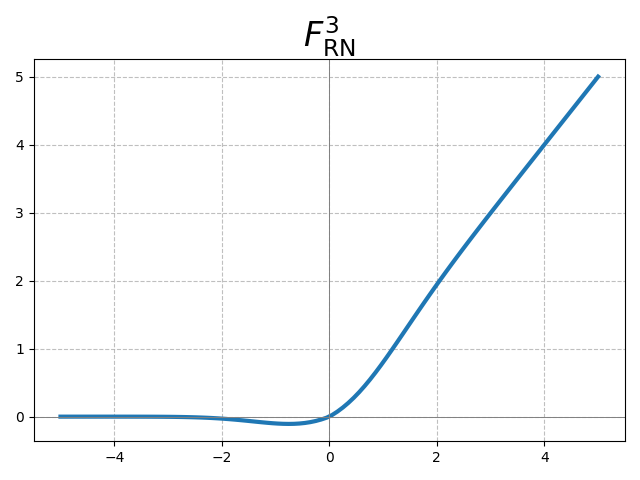}
\end{subfigure}
\begin{subfigure}{0.32\textwidth}
\centering
	\includegraphics[width=\linewidth]{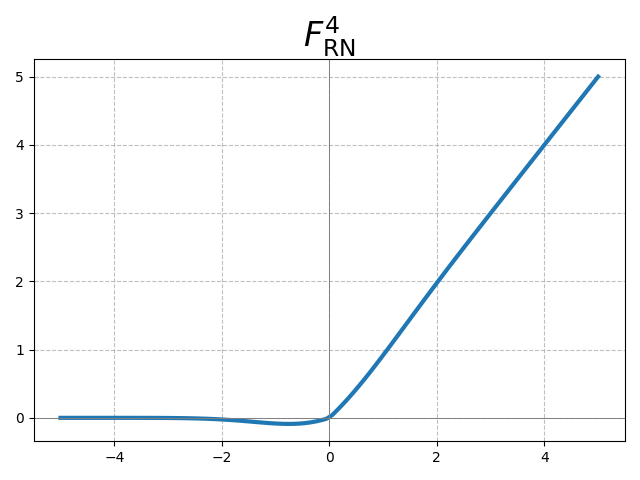}
\end{subfigure}
\begin{subfigure}{0.32\textwidth}
\centering
	\includegraphics[width=\linewidth]{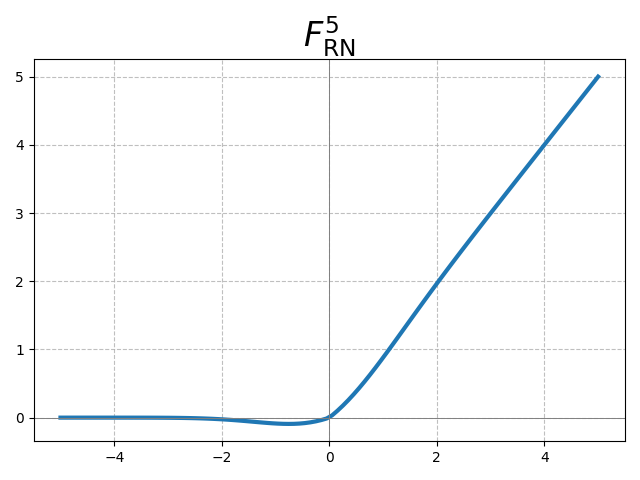}
\end{subfigure}
\caption{Plots of activation functions in Eq.\ref{eq:rn}, found on ResNet20 / CIFAR10.}
	\label{plot:rn}
\end{figure}

\begin{figure}[H]
\centering
\begin{subfigure}{0.32\textwidth}
\centering
	\includegraphics[width=\linewidth]{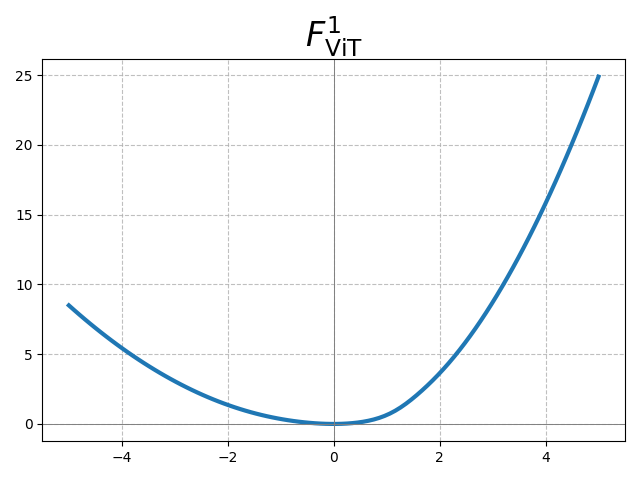}
\end{subfigure}
\begin{subfigure}{0.32\textwidth}
\centering
	\includegraphics[width=\linewidth]{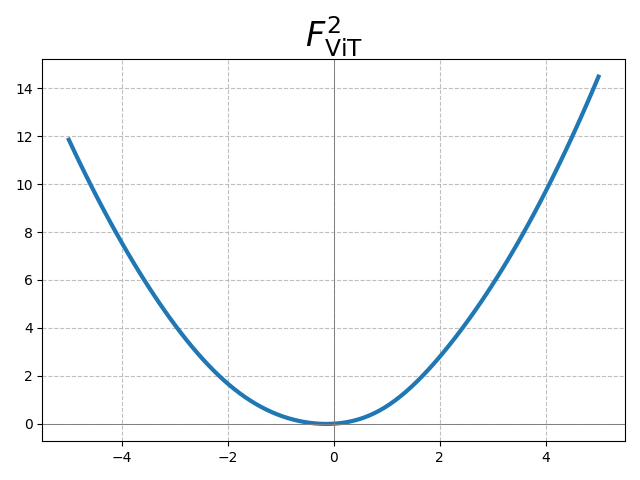}
\end{subfigure}
\begin{subfigure}{0.32\textwidth}
\centering
	\includegraphics[width=\linewidth]{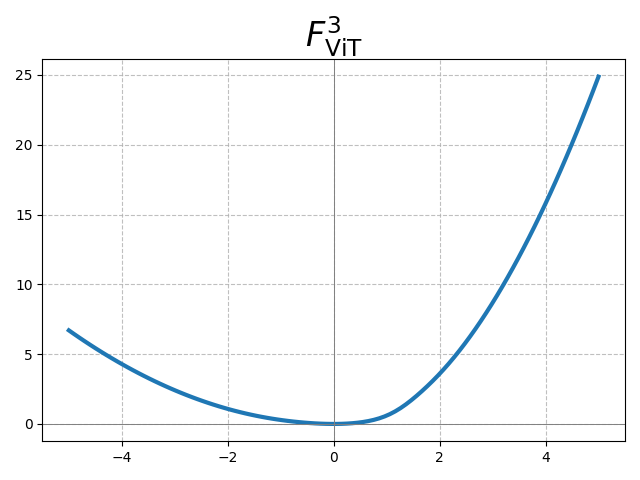}
\end{subfigure}
\begin{subfigure}{0.32\textwidth}
\centering
	\includegraphics[width=\linewidth]{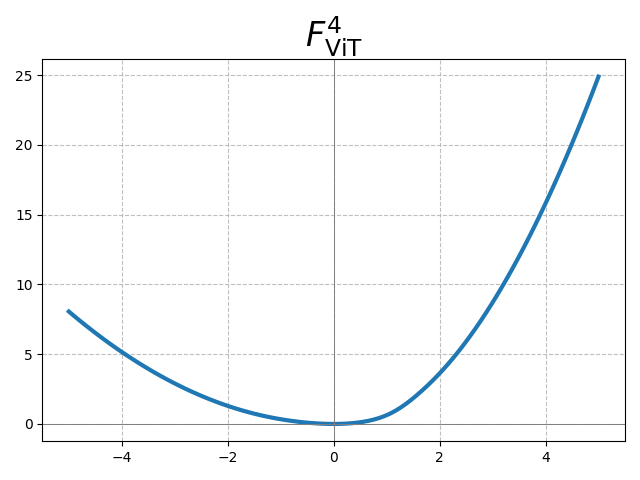}
\end{subfigure}
\begin{subfigure}{0.32\textwidth}
\centering
	\includegraphics[width=\linewidth]{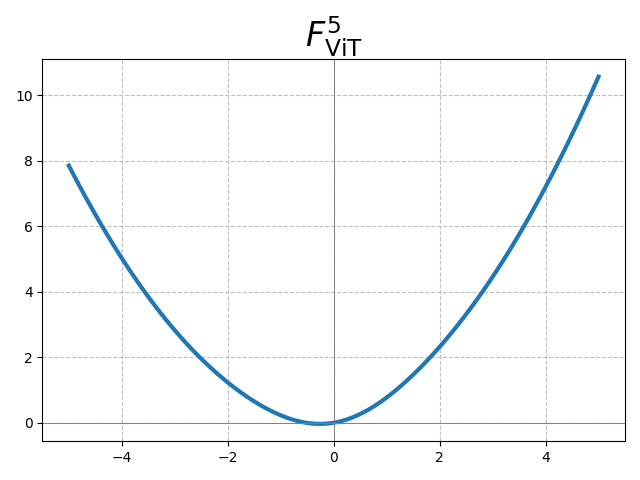}
\end{subfigure}
	\caption{Plots of activation functions in Eq.\ref{eq:vit}, found on ViT-Tiny / CIFAR10.}
	\label{plot:vit}
\end{figure}

\begin{figure}[H]
\centering
\begin{subfigure}{0.32\textwidth}
\centering
	\includegraphics[width=\linewidth]{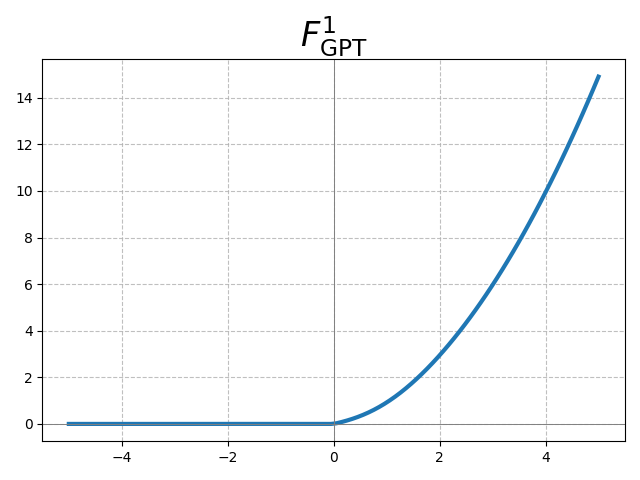}
\end{subfigure}
\begin{subfigure}{0.32\textwidth}
\centering
	\includegraphics[width=\linewidth]{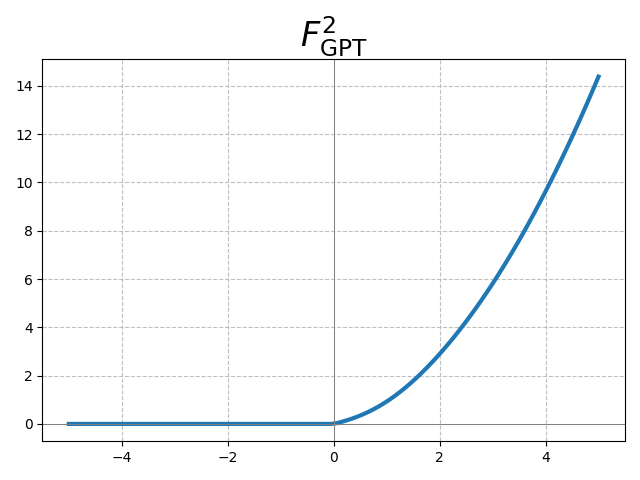}
\end{subfigure}
\begin{subfigure}{0.32\textwidth}
\centering
	\includegraphics[width=\linewidth]{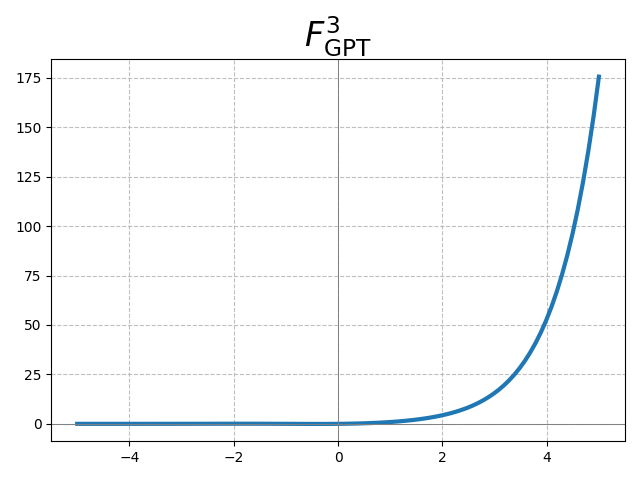}
\end{subfigure}
\begin{subfigure}{0.32\textwidth}
\centering
	\includegraphics[width=\linewidth]{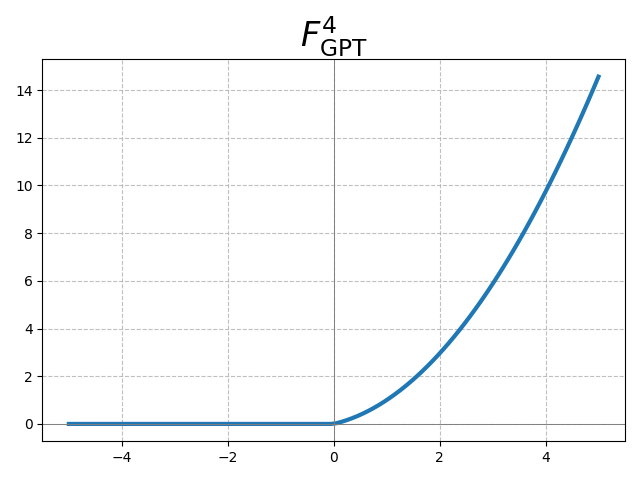}
\end{subfigure}
\begin{subfigure}{0.32\textwidth}
\centering
	\includegraphics[width=\linewidth]{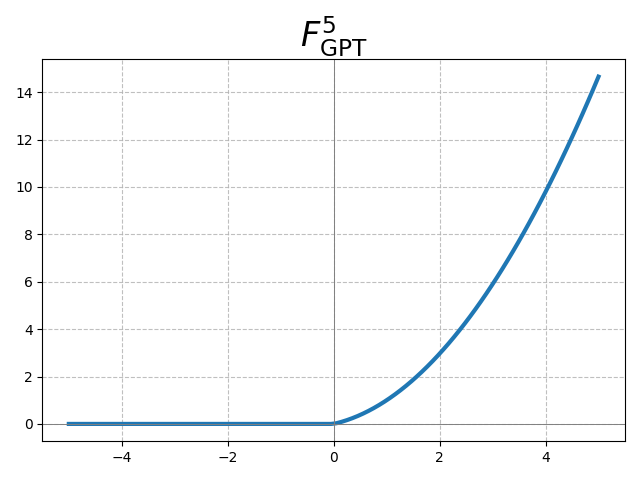}
\end{subfigure}
	\caption{Plots of activation functions in Eq.\ref{eq:gpt} found on miniGPT / TinyStories.}
	\label{plot:gpt}
\end{figure}

\end{document}